\documentclass{article}
\usepackage{ijcai21}

\usepackage{times}
\usepackage{soul}
\usepackage{url}
\usepackage[hidelinks]{hyperref}
\usepackage[utf8]{inputenc}
\usepackage[small]{caption}
\usepackage{graphicx}
\usepackage{amsmath}
\usepackage{amsthm}
\usepackage{booktabs}
\usepackage{algorithm}
\usepackage{algorithmic}
\title{Contextualized Embeddings based Convolutional Neural Networks for Duplicate Question Identification}

\author{
    Content Areas:
    \affiliations
    Paraphrase Identification \\
    Semantic Equivalence \\
    Transformer Architecture
}

\author{
Harsh Sakhrani$^1$\thanks{Equal Contributors}
\and
Saloni Parekh$^1$\footnotemark[1]\And
Pratik Ratadiya$^2$
\affiliations
$^1$Pune Institute of Computer Technology, Maharashtra, India\\
$^2$vCreaTek Consulting Services Pvt. Ltd., Maharashtra, India\\
\emails
\{harshsakhrani26, saloniparekh1609\}@gmail.com,
pratik.r@vcreatek.com
}

\begin{document}

\maketitle

\begin{abstract}
  Question Paraphrase Identification (QPI) is a critical task for large-scale Question-Answering forums. The purpose of QPI is to determine whether a given pair of questions are semantically identical or not. Previous approaches for this task have yielded promising results, but have often relied on complex recurrence mechanisms that are expensive and time-consuming in nature. In this paper, we propose a novel architecture combining a Bidirectional Transformer Encoder with Convolutional Neural Networks for the QPI task. We produce the predictions from the proposed architecture using two different inference setups: Siamese and Matched Aggregation. Experimental results demonstrate that our model achieves state-of-the-art performance on the Quora Question Pairs dataset. We empirically prove that the addition of convolution layers to the model architecture improves the results in both inference setups. We also investigate the impact of partial and complete fine-tuning and analyze the trade-off between computational power and accuracy in the process. Based on the obtained results, we conclude that the Matched-Aggregation setup consistently outperforms the Siamese setup. Our work provides insights into what architecture combinations and setups are likely to produce better results for the QPI task.
\end{abstract}

\section{Introduction}
\par Paraphrase Identification is a task that aims to recognize whether two input sequences are semantically equivalent or not, and is a binary classification problem. Question Paraphrase Identification is a variation of Paraphrase Identification where the two sequences are interrogative in nature. It is a critical task for Question-Answering forums like Quora~\footnote{https://www.quora.com/}, which can benefit from merging questions with similar meanings and organizing them more efficiently. Moreover, it can also help in the retrieval of questions that are semantically equivalent to a question asked by the user.
\par Deep Learning-based mechanisms for Paraphrase Identification can be classified into two major frameworks. The first framework is the Siamese Architecture that has a common encoder that is applied separately to both the input sentences to generate sentence encodings in the same embedding space ~\cite{yu2014deep,bowman2016fast}. The framework has the advantage of being lightweight and simple to train because of shared network parameters. However, the disadvantage is that there is no explicit interaction between the two sentences in the encoding process. Some methods suggest the use of an attention mechanism to improve this interaction~\cite{tan-etal-2016-improved}. Such a method uses one sequence's representation to attend to another but fails to capture the word-level interactions.
\par To overcome this disadvantage, a second framework called Matched Aggregation has been proposed ~\cite{wang2016compare,wang2017bilateral}. In this framework, the attention mechanism is utilized at the word level to get the matching words between the two sentences. The matching information is then aggregated into a vector to make the final classification decision. This framework allows for word-level interaction, ensuring significantly improved results. However, most of the previous Matched Aggregation approaches relied on a combination of recurrence and complex attention ensembles to match the two sentences at the word level~\cite{lan-xu-2018-neural}.
\par To this end, we propose a contextual embedding-based convolution architecture for question paraphrase or duplicate question identification. We propose the use of the Bidirectional Transformer Encoder, based on~\cite{devlin2018bert}, that is capable of generating contextual word embeddings utilizing the self-attention mechanism. In addition to this, we also make use of Convolutional Neural Networks ~\cite{kim-2014-convolutional} to extract the semantic level features from the trained contextual embeddings. We perform extensive experiments utilizing the proposed architecture in both setups- Siamese and Matched Aggregation. The system is evaluated on the Quora Question Pairs Dataset ~\footnote{https://www.kaggle.com/c/quora-question-pairs} and produces state-of-the-art results.
\par Specifically, we make the following contributions through this paper: 
\begin{itemize}
    \item We present a new approach for question paraphrase identification in two different inference setups - Siamese and Matched Aggregation, leveraging the Bidirectional Transformer Encoder’s pretraining on the Next Sentence Prediction task.
    \item We achieve state-of-the-art results on the Quora Question pairs dataset on which we prove that the addition of convolutional layers helps to improve the performance as against using a vanilla transformer architecture.
    \item Furthermore, we also experiment with partial and complete fine-tuning and empirically show that one doesn’t need to increase the trainable parameters beyond a certain extent to achieve improved results, thereby also assessing the trade-off between computational power and accuracy.
\end{itemize}
\par The rest of the paper is structured as follows: Section 2 talks about the previous approaches which have been used to tackle the problem, Section 3 explains the proposed methodology in detail, Section 4 talks about the description of the dataset, Section 5 presents the results and analysis of the proposed system against previous approaches while Section 6 concludes our paper.

\section{Background Work}
\par The paraphrase identification task, a well-studied Natural Language Processing (NLP) problem, uses Natural Language Sentence Matching (NLSM) to decide whether two sentences are paraphrased or not. Identifying duplicate questions can be thought of as a special use case of this task.
\par Over time, researchers have provided multiple approaches to solve this problem by trying to break down the complex internal structures of sentences and the interactions between them. The two of the most used approaches are mentioned in this section. The first method employs a "Siamese" setup, which is made up of two sub-networks with common parameters. Each sub-network responds to a single sentence and the outputs corresponding to them are mathematically operated on to obtain the final result.
\par ~\cite{chen2018quora} used the Manhattan LSTM model and merged the outputs of the two subnetworks for the final classification decision.~\cite{brahma2018regmapr} augmented the embeddings obtained from a Siamese architecture with two features, based on exact and paraphrase match between the words in two sentences.~\cite{godbole2018siamese} combined the features from Gated Recurrent Unit sub-networks with additional hand-crafted feature representations and used machine learning algorithms like Random Forest Classifier and Support Vector Machines (SVM) to obtain the results.~\cite{chopra2020applying} used a combination of similarity scores obtained from a Siamese network over LSTM and three other scores (two variants of cosine similarity and customized fuzzy match) and developed a meta classifier using SVMs. 
\par The second setup, Matched Aggregation understands the relationship between the two sentences at the word level. Usually, the solutions in this space resort to attention to aggregate the matching information.~\cite{wang2017bilateral} proposed the Bilateral Multi-Perspective Matching (BiMPM) model that matches two sentences by a “bilateral matching with attention” mechanism in multiple perspectives. They proposed four types of representations instead of the attention-weighted representation to improve the results.~\cite{tomar2017neural} used character n-grams instead of word embeddings to improve the performance of BiMPM and thus matched the sentences from multiple perspectives.~\cite{gong2017natural} introduced the Interactive Inference Network (IIN), which can hierarchically extract semantic features from the interaction space.~\cite{ijcai2018-613} applied 4 different attention mechanisms on the obtained contextual embeddings and aggregated the matching information by further making use of bi-directional RNNs.~\cite{choi2019cell} constructed multi-layer LSTMs where memory cell states from the previous layer are used to control the vertical information flow thus filtering the information from lower layers and reflecting it through a soft gating mechanism.~\cite{XU2020113384} used an enhanced attention mechanism that helped strengthen the interaction between sentences via adding alignment context into local context in the convolution operation and combining multi-grained similarity features in different filter sizes.~\cite{mfae} used BERT to obtain the dynamic pre-trained word embeddings. Then inter and intra-asking emphasis is obtained by summing inter-attention and self-attention, respectively. The idea is that, the more a word interacts with others, the more important the word is. Finally, an eight-way combination is used to generate multi-fusion asking emphasis and multi-fusion word representation.
\par These approaches, however, have some drawbacks of their own. In the Siamese setting, there is no interaction between the two sentences in the encoding process. On the other hand, the Matched Aggregation approach makes use of recurrence in the encoding process and complex attention mechanisms to improve the word-level interaction between the two sentences. Some methods which use this approach perform matching only in a single direction, neglecting the information in the sentence pairs. We aim to tackle these drawbacks by employing a combination of transformer architecture and CNNs that provides for bidirectionality and context between the two phrases while relying just on attention and convolutional operations. Transfer learning has been on the rise in NLP ~\cite{malte2019evolution}, and we intend to leverage the same for our benefit.

\section{Proposed Methodology}
\par In this section, we explain the proposed model architecture. First, we give a brief description of the Question Paraphrase Identification Task, followed by definitions of the building blocks of the proposed architecture. This is followed by the two inference setups and the hyperparameter setup for our experiment.

\subsection{Task Description}
\par For the question paraphrase identification task, formally, each question pair can be represented as a triplet ($A$,$B$,$y$), where $A$ = ($a_1$ ,..., $a_i$ ,..., $a_N$) is a question sequence of length $N$, $B$ = ($b_1$ ,...,$ b_j$ ,..., $b_M$) is a question sequence of length $M$, and $y$ $\in$ $\{0, 1\}$ is the label that represents the relationship between $A$ and $B$. Here, $y$ = $1$ denotes that $A$ and $B$ are identical, while $y$ = $0$ denotes that they are semantically dissimilar.

\subsection{Architecture Details}
\par We propose a combination of the Bidirectional Transformer Encoder and Convolutional Neural Networks as the main feature extraction block. Contextual embeddings are derived from the transformer encoder which are then passed as an input to the convolutional network. 

\subsubsection{Bidirectional Transformer Encoder (BTE)}
\par
\begin{figure}[t]
    \centering
    \includegraphics[width=7cm]{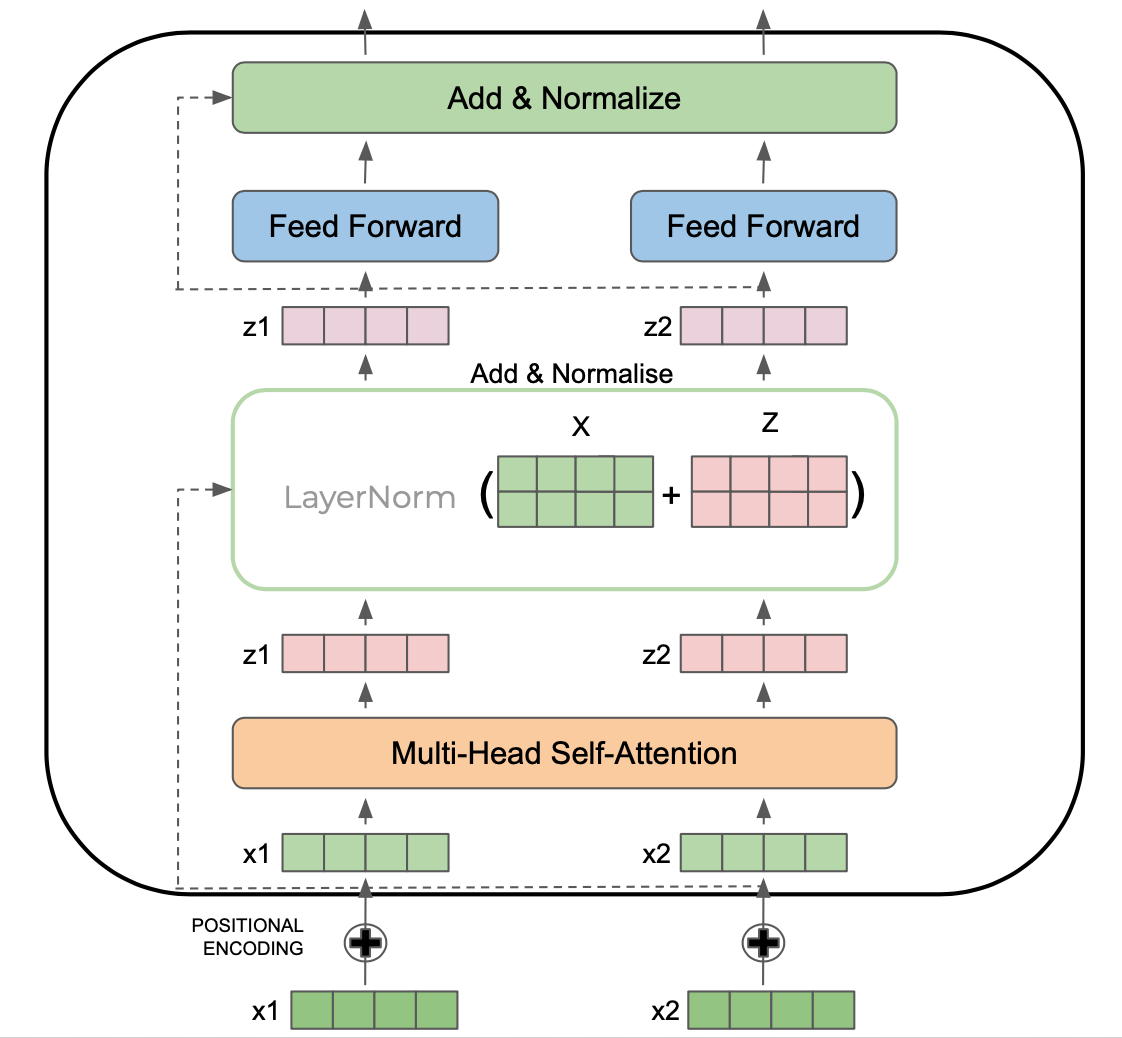}
    \caption{Bidirectional Transformer Encoder}
    \label{fig:encoder}
\end{figure}

 The transformer encoder block is inspired from the BERT model architecture ~\cite{devlin2018bert}. The internal architecture is significantly dominated by the Transformer encoder ~\cite{vaswani2017attention}. The model was pre-trained on large text corpora and can generate word-level contextualized representations for a given sequence. As shown in Figure~\ref{fig:encoder}, the encoder block comprises a Multi-Head Self Attention Layer and a position-wise fully connected feed-forward layer. A residual connection is employed between each of these two layers, followed by layer normalization. 
\par The encoder employs the “Scaled Dot Product Attention” mechanism to encode each token based on all the other relevant tokens in the sequence. The formula for calculating attention focused weights is shown in the following equation:

\begin{equation}
    Z = softmax ( \frac{Q \times K^T}{\sqrt{(d_k)}}) V
\end{equation}
\par where, $Q$ = ($W_qE$) is the Query vector, $K$ = ($W_kE$) is the Key vector, $V$ = ($W_vE$) is the Value vector. Here $E$ is the embedding vector and $W_q$, $W_k$ and $W_v$ are trainable Weight matrices.
\par The attention mechanism is further refined by employing “Multi Headed” attention. The idea is to leverage the attention mechanism multiple times by calculating multiple $Z$ matrices for different sets of Key, Query, and Value vectors. All the output $Z$ matrices are then concatenated and further multiplied by an additional weight matrix to get the corresponding Multi Headed Attention output which is then fed to the feed-forward layer.
\par We use the publicly available weights of the Bidirectional Transformer Encoder\footnote{https://huggingface.co/bert-base-uncased}, which has \textbf{12} Transformer Encoders, \textbf{12} self-attention heads and an embedding size of \textbf{768}. 

\begin{figure}[t]
    \centering
    \includegraphics[width=8cm]{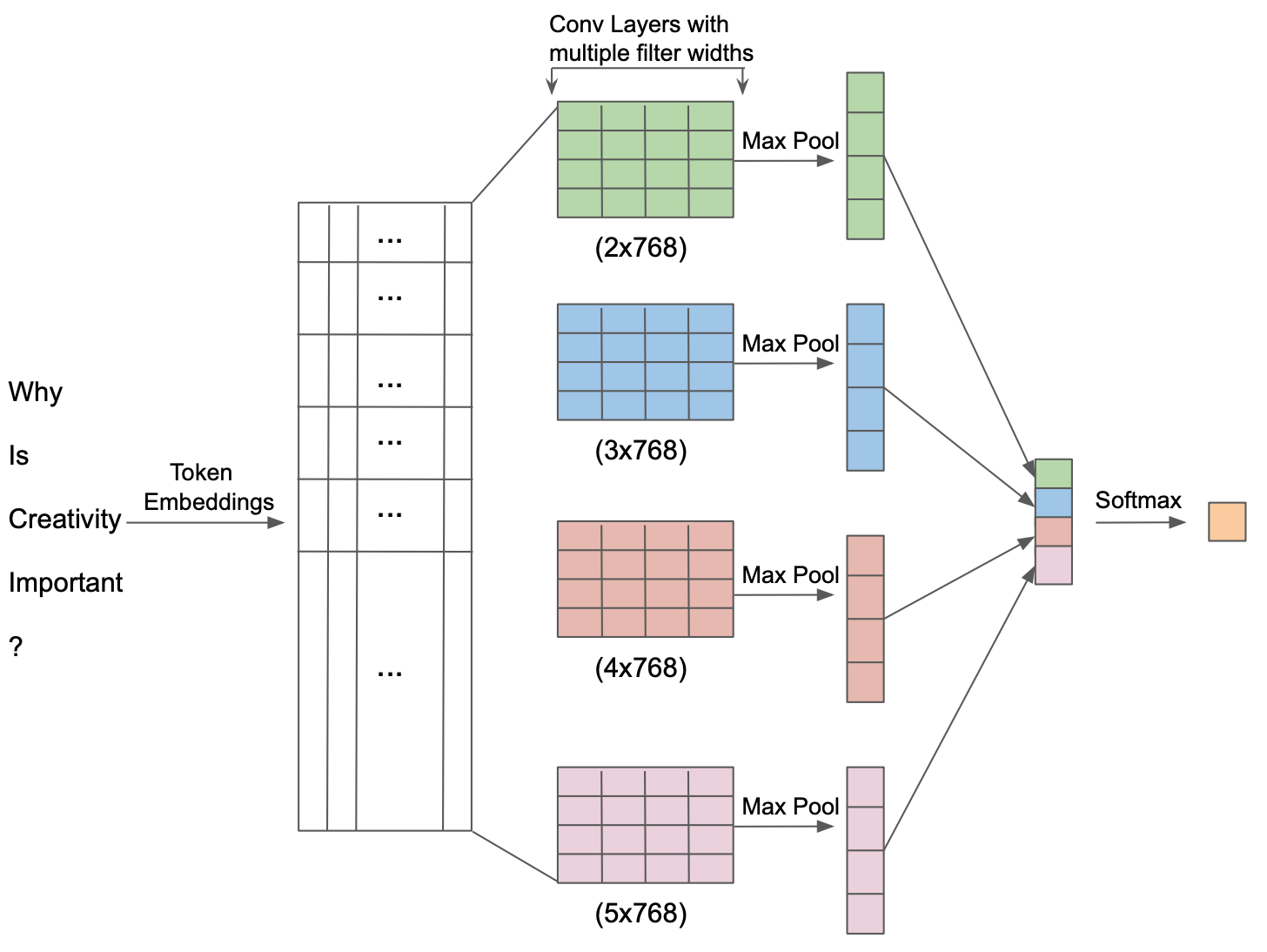}
    \caption{CNN Model Architecture}
    \label{fig:kimcnn}
\end{figure}

\begin{figure*}[t]
    \centering
    \includegraphics[width=7in]{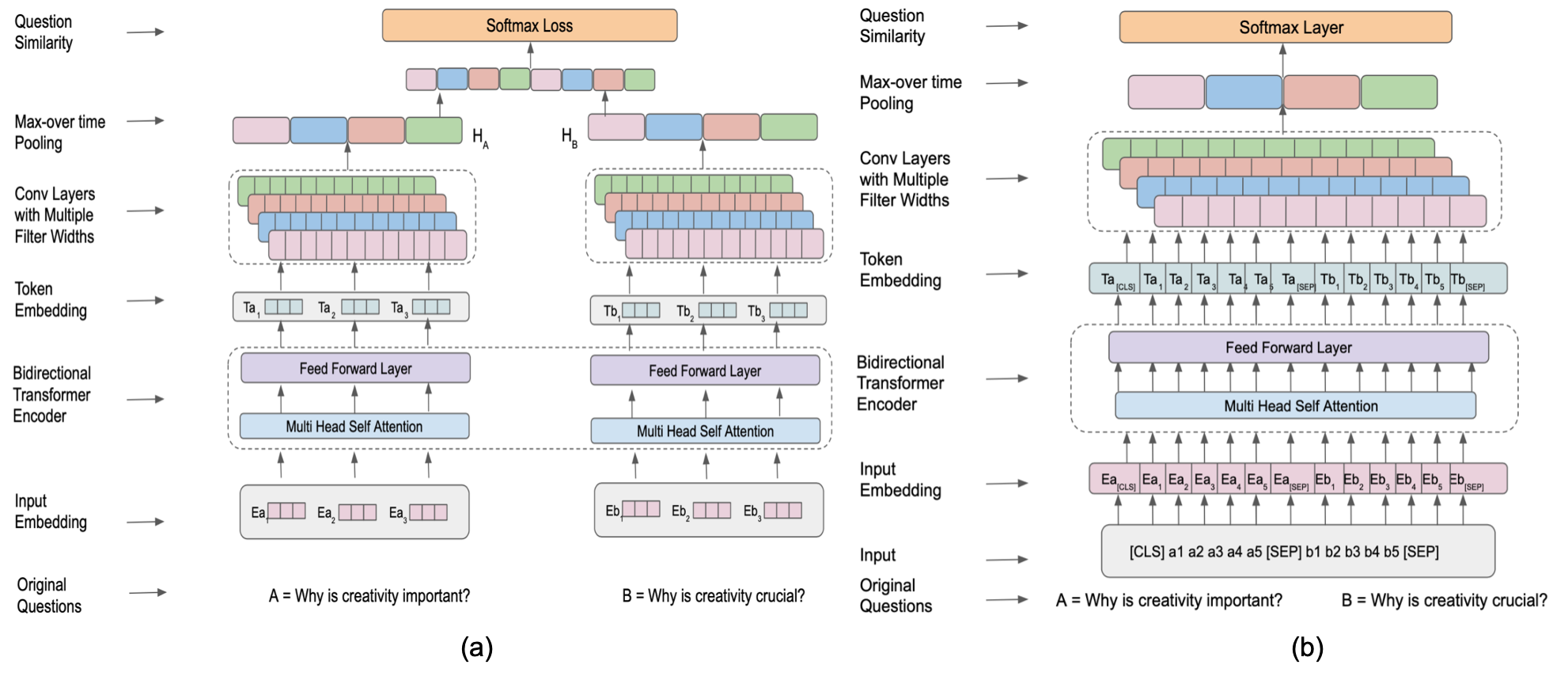}
    \caption{Our Proposed Methodology - a combination of Bidirectional Transformer Encoder and CNN: (a) Siamese Framework. (b) Matched Aggregation Framework.}
    \label{fig:architecture}
\end{figure*}

\subsubsection{Convolutional Neural Networks (CNN)}
\par The CNN architecture we adopt is shown in Figure~\ref{fig:kimcnn}. Let $x_i$ $\in$ $R^{768}$ be the token embedding of the $i^{th}$ token in the sequence. The resulting output representations from the encoder (of length $n$) can be interpreted as:
\begin{equation}
    x_{1:n} = x_1 \oplus x_2 \oplus ... \oplus x_n
\end{equation}
\par where $\oplus$ is the concatenation operator. The convolution operation applies a filter $w$ $\in$ $R^{g,768}$ to a $g$-word window to produce a new feature. A feature $e_i$ is generated from a window of words $x_{i:i+g-1}$ by:
\begin{equation}
    e_i = f(w \cdot x_{i:i+g-1} + b)
\end{equation}
\par where $b$ is the bias term and $f$ is the ReLU activation function. This filter is then applied to each possible window of words in the sequence \{$x_{1:g}$, $x_{2:g+1}$, ... , $x_{n-g+1:n}$\} to produce a feature map:
\begin{equation}
    e = [e_1, e_2, ... , e_{n-g+1}]
\end{equation}
\par The feature map is then subjected to a max-over-time pooling operation, with the maximum value serving as the feature corresponding to a particular filter. 
\par This process is for extracting a single feature using a single filter. To obtain multiple features, the model employs multiple filters (of varying window sizes). The CNN in our case is made up of $400$ parallel convolutional filters of four distinct sizes $(768\times2, 768\times3, 768\times4, 768\times5)$, each with $100$ filters. These features are then concatenated and passed through a fully connected softmax layer to obtain the corresponding label. We also adopt a dropout layer for regularization purposes.

\subsection{Inference Methodology}
As mentioned earlier, we experiment with two inference setups for the proposed architecture:
\begin{enumerate}
    \item Siamese Network 
    \item Matched Aggregation Framework
\end{enumerate}
\par In the case of Siamese Architecture, as shown in Figure~\ref{fig:architecture}(a), our proposed approach works in the following steps: The input question is first tokenized, and an input embedding is created for each token $a_i$ $\in$ $A$ and $b_i$ $\in$ $B$. These embeddings are then fed to the Encoder along with their corresponding attention masks. The attention mask is a boolean vector that represents the tokens that should or should not be attended to by the model. The resulting output representations from the encoder are then passed through multiple parallel convolutional filters and max-pooling layers to obtain multiple output features. Concatenating these multiple output features yields condensed vectors $H_A$ and $H_B$, which represent questions $A$ and $B$, respectively. The condensed vectors are then concatenated and passed through a softmax layer to get the corresponding label. Our proposed methodology for the Siamese Network is explained in Algorithm~\ref{alg:algorithm1}.

\begin{algorithm}
\caption{Siamese Network}
\label{alg:algorithm1}
\textbf{Input}: $T_n$ tuples, each of form ($S_a$, $S_b$, $label$);\\
\textbf{Output}: Trained model
\begin{algorithmic}[1] 
\FOR{$tuple$ $\in$ $T_n$}
\STATE $h_a$ $\leftarrow$ $BTE+CNN$($tuple$[0])
\STATE $h_b$ $\leftarrow$ $BTE+CNN$($tuple$[1])
\STATE $label$ $\leftarrow$ $tuple$[2]
\STATE $input$ $\leftarrow$ $h_a$ $\oplus$ $h_b$
\STATE $output$ $\leftarrow$ softmax($input$)
\STATE backpropogate(CROSS\_ENTROPY\_LOSS($output$,$label$))
\ENDFOR
\end{algorithmic}
\end{algorithm}

\par In the Matched Aggregation approach as shown in Figure~\ref{fig:architecture}(b), the two questions are “packed together” into a single sequence. The questions are differentiated by the presence of a special token ($[SEP]$) between the two questions. Furthermore, a binary mask is also associated with every token indicating whether it belongs to question $A$ or question $B$. Since the two questions are packed into a single sequence, one single condensed vector is obtained (corresponding to the sequence) which is then passed through a softmax layer to get the corresponding label. Our proposed methodology for the Matched Aggregation Framework is explained in Algorithm~\ref{alg:algorithm2}. The difference in the encoding process is the significant distinguishing factor between the two proposed approaches.

\begin{algorithm}
\caption{Matched Aggregation Framework}
\label{alg:algorithm2}
\textbf{Input}: $T_n$ tuples of form ($S_a$, $S_b$, $label$);\\
\textbf{Output}: Trained model
\begin{algorithmic}[1] 
\FOR{$tuple$ $\in$ $T_n$}
\STATE $inp$ $\leftarrow$ [CLS] + $tuple$[0] + [SEP] + $tuple$[1] + [SEP]
\STATE $label$ $\leftarrow$ $tuple$[2]
\STATE $h_{int}$ $\leftarrow$ $BTE+CNN$($inp$)
\STATE $output$ $\leftarrow$ softmax($h_{int}$)
\STATE backpropogate(CROSS\_ENTROPY\_LOSS($output$,$label$))
\ENDFOR
\end{algorithmic}
\end{algorithm}

\subsection{Hyperparameter Setup}
\par The hyperparameter setup is identical for both the Siamese Network and the Matched Aggregation Framework except for the \textit{max\_len} argument. The \textit{max\_len} of the input sequence was chosen to be \textbf{64} for the Matched Aggregation Framework and \textbf{32} for the Siamese Framework. To counter the slight imbalance in the dataset, we adopt the \textit{Weighted Random Sampler} from PyTorch. We use the Cross-Entropy Loss function to calculate the loss. Adam was used as an optimizer, the batch size was chosen to be \textbf{8}, and the learning rate was set to $10^{-5}$. The models were trained for \textbf{12} epochs on Nvidia GeForce RTX 2080 Ti GPU.

\begin{table}[t]
\centering
\begin{tabular}{ll}
\toprule
\textbf{Identification of Duplicate Questions}\\
\midrule
\textbf{$S_1$}: How can I be a good geologist? \\
\textbf{$S_2$}: What should I do to be a great geologist? \\
\textbf{Label}: 1 (Duplicate Questions) \\
\textbf{$S_1$}: What are some good rap songs to dance to? \\
\textbf{$S_2$}: What are some of the best rap songs? \\
\textbf{Label}: 0 (Distinct Questions)\\
\bottomrule
\end{tabular}
\caption{Sample dataset inputs}
\label{tab:dataset_ex}
\end{table}

\section{Dataset Description}
The Quora Question Pairs dataset comprises of more than 400,000 question pairs along with their indicative paraphrase label. There is no official train/validation/test split. However, a standard split used by~\cite{wang2017bilateral} has since been used by other studies as well~\cite{tomar2017neural,gong2017natural,ijcai2018-613}. We use the same split for fair result comparison. Both the validation and test set comprise of 10000 samples each, 5000 paraphrase pairs, and 5000 non-paraphrase pairs. Table~\ref{tab:dataset_ex} shows some sample inputs from the dataset. 


\section{Results}
\par For the comparison against previous techniques, we report the Classification accuracy in Table~\ref{tab:exp2} for duplicate question identification on the Quora Question Pairs Test Set. Due to the fair balance of samples between the classes, most papers utilize test accuracy as the only evaluation metric. It can be seen that our BTE+CNN approach in a matched aggregation framework has produced the best results.
\par We also conduct experiments to determine how the number of tunable encoders impacts the overall performance of the model. Experimental results suggest that there is an increase in the performance with the number of tunable encoders, however, it is not substantial enough to compensate for the massive increase in the training cost and time. We also carry out experiments to analyze the influence of CNN on the model architecture and the overall performance. When compared to Mean Pooling, the incorporation of CNN in the design results in considerably superior performance. We also report the F1 scores for the experiments we conducted to provide a more comprehensive understanding of our methodology. All these results are tabulated in Table ~\ref{tab:exp1}.

\begin{table}[t]
\centering
\begin{tabular}{lr}
\toprule
Model  & Accuracy (\%)\\
\midrule
Siamese-LSTM~\cite{wang2017bilateral} &82.58 \\
MP-LSTM~\cite{wang2017bilateral} &83.21 \\
L.D.C.~\cite{wang-etal-2016-sentence} &85.55 \\
BiMPM~\cite{wang2017bilateral} &88.17 \\
\midrule
PWIM~\cite{he-lin-2016-pairwise} &83.40 \\
ESIM~\cite{chen-etal-2017-enhanced} &85.00 \\
ESIM+Syn T.LSTM~\cite{chen-etal-2017-enhanced} &85.40 \\
InferSent~\cite{conneau2017supervised} &86.60 \\
SSE~\cite{nie2017shortcut} &87.80 \\
\midrule
GenSen~\cite{subramanian2018learning} &87.01 \\
LSTM+ElBiS~\cite{choi2018element} &87.30 \\
CDTFME-Aver~\cite{xie2019dynamic} &88.00 \\
pt-DecAttword~\cite{tomar2017neural} &87.54 \\
pt-DecAttchar~\cite{tomar2017neural} &88.40 \\
CAS-LSTM~\cite{choi2019cell} &88.40 \\
Bi-CAS-LSTM~\cite{choi2019cell} &88.60 \\
REGMAPR~\cite{brahma2018regmapr} &88.64 \\
DIIN~\cite{gong2017natural} &89.06 \\
MwAN\cite{ijcai2018-613} &89.12 \\
DIIN (Ensemble)\cite{gong2017natural} &89.84 \\
MFAE(ELMo)~\cite{mfae} &89.61 \\
MFAE(BERT)~\cite{mfae} &89.79 \\
MFAE(BERT Ens)~\cite{mfae} &90.54 \\
\midrule
\textbf{BTE + CNN (4 trainable encoders)\footnotemark} &\textbf{90.58} \\
\textbf{BTE + Mean Pooling (12 trainable encoders)\footnotemark[\value{footnote}]}   &\textbf{90.78} \\
\textbf{BTE + CNN (12 trainable encoders)\footnotemark[\value{footnote}]} &\textbf{90.80} \\
\bottomrule
\end{tabular}
\caption{Classification Accuracy for Duplicate Question Identification on Quora Question Pairs}
\label{tab:exp2}
\end{table}

\footnotetext{This model belongs to the Matched Aggregation Framework}

\begin{table*}[t]
\centering
\begin{tabular}{llrrrr}
\toprule
Setup &Model &No. of Tr. Encoders  & Accuracy (\%) & F1-Score &No. of Tr. Params\\
\midrule
Siamese &BTE+Mean Pooling   &4    &81.11  &0.8008 &29M \\
        &BTE+Mean Pooling                   &12   &82.10  &0.8085 &109M \\
        &BTE+CNN           &2    &86.68  &0.8606 &16M \\
        &BTE+CNN                  &4    &\textbf{86.88}  &\textbf{0.8613} &30M \\
        &BTE+CNN                  &12    &86.56  &0.8562 &110M \\
        
\midrule
Matched Aggregation     &BTE+Mean Pooling   &4   &90.41 &0.8981 &29M \\
                        &BTE+Mean Pooling                   &12  &90.78 &0.9027 &109M \\
                        &BTE+CNN            &2   &90.10 &0.8950 &16M \\
                        &BTE+CNN                   &4   &90.58 &0.9002 &30M \\
                        &BTE+CNN                   &12   &\textbf{90.80} &\textbf{0.9022} &110M \\
\bottomrule
\end{tabular}
\caption{Performance Comparison in the two setups - Siamese and Matched Aggregation. Tr stands for Trainable.}
\label{tab:exp1}
\end{table*}

\begin{table}[t]
\centering
\begin{tabular}{ll}
\toprule
\textbf{Predictions by the MA and Siamese Frameworks}\\
\midrule
\textbf{$S_1$}: How do I improve improvisation skills on drums?  \\
\textbf{$S_2$}: What should I do to improve my drumming skills?  \\
\textbf{MA}: 0 (Distinct Questions) \\
\textbf{Siam}: 1 (Duplicate Questions)\\
\bottomrule
\end{tabular}
\caption{Predictions. MA: Matched Aggregation Framework and Siam: Siamese Framework}
\label{tab:pred_ex}
\end{table}

\par Our main observations after performing this study have been:
\begin{enumerate}
    \item The model can recognize stronger relationships between the two sentences due to the Bidirectional Transformer Encoder's pre-training on the Next Sentence Prediction task and its capacity to pack two sentences together. This differs from the Siamese network approach in which the two sentences do not interact explicitly during the encoding process. 
    \item CNNs seem to be more effective than mean pooling because their design enhances the quality of feature representations by introducing more semantic relationships. In the instance of the Siamese setup, the combination of the Bidirectional Transformer Encoder and CNNs with just four trainable encoders outperforms a fully fine-tuned Bidirectional Transformer Encoder alone. This demonstrates how CNN aids the Siamese framework in capturing the information.
    \item Experimenting with a different number of trainable parameters suggests that the higher the number of trainable parameters, the sharper the performance would be. For instance, in the case of Matched Aggregation Framework, the Bidirectional Transformer Encoder-CNN hybrid that outperforms previous state-of-the-art baselines has only four trainable encoders. Even though slightly better results were yielded by training Bidirectional Transformer Encoder in its entirety (twelve trainable encoders), it comes at the cost of having $\sim$4x more trainable parameters, introducing a huge computational overload.
    \item We consider the BTE-CNN hybrid with 12 encoders in both the frameworks and discover that the Matched Aggregation framework was able to correctly recognize 62\% of the samples that were incorrectly classified by the Siamese framework on the test set. Table~\ref{tab:pred_ex} shows a sample that was wrongly classified by the Siamese Framework as a duplicate pair, but correctly classified by the Matched Aggregation Framework as a distinct pair. 
\end{enumerate}

\section{Conclusion}
In this paper, we introduced a novel contextual embedding-based CNN architecture for duplicate question pairs identification on the Quora Question Pairs dataset. We combined together Bidirectional Transformer Encoder and Convolutional Neural Networks and leveraged its representation extraction ability in two setups - Siamese and Matched Aggregation. More importantly, we discover that our approach of encoding the two sentences together proves to be effective in identifying the duplication. Results show that the proposed approach produces the best results yet, despite being a non-ensemble approach with significantly less training cost. Another point to note is that, while training the Encoder in its entirety ensures even better results, the current trained setup almost matches the same results while having $\sim$4x lesser trainable parameters. The future direction of this work could include investigating the performance of the model on multilingual and domain-specific data.

\bibliographystyle{named}
\bibliography{ijcai21}

\end{document}